\documentclass[10pt,twocolumn,letterpaper]{article}

\usepackage{cvpr}              
\definecolor{cvprblue}{rgb}{0.21,0.49,0.74}
\usepackage[pagebackref,breaklinks,colorlinks,allcolors=cvprblue]{hyperref}
\usepackage{multirow}
\usepackage[accsupp]{axessibility}

\def\paperID{11188} 
\def\confName{CVPR}
\def\confYear{2026}

\title{PointCSP: Cross-Sample Semantic Propagation and Stability Preservation in Self-Supervised Point Cloud Learning}

\author{
    Xinxing Yu\textsuperscript{\rm 1}, 
    Ajian Liu\textsuperscript{\rm 1}\textsuperscript{\rm 3}\thanks{Corresponding author}, 
    Sunyuan Qiang\textsuperscript{\rm 2}, 
    Hui Ma\textsuperscript{\rm 1}\textsuperscript{\rm 4}\thanks{With the assistance of Hui Ma's writing guidance.}, 
    Liying Yang\textsuperscript{\rm 1}, \\
    Yuzhong Wang\textsuperscript{\rm 1}, 
    Zhi Rao\textsuperscript{\rm 1}, 
    Yanyan Liang\textsuperscript{\rm 1}$^{\ast}$\\
    \textsuperscript{\rm 1}Faculty of Innovation Engineering, Macau University of Science and Technology, Macau, China\\
    \textsuperscript{\rm 2}Southwest Institute of Technical Physics, Chengdu, China\\
    \textsuperscript{\rm 3}MAIS, The Institute of Automation of the Chinese Academy of Sciences, Beijing, China\\
    \textsuperscript{\rm 4}School of Computing and Information Technology, Great Bay University, Dongguan, China\\
    {\tt\small \{astrumyu, mahuilight, lyyang69, yzwangjoseph, rz1445692095\}@gmail.com}\\
    {\tt\small {qiangsunyuan2025@163.com, ajian.liu@ia.ac.cn, yyliang@must.edu.mo}}
}

\begin{document}
\maketitle
\begin{abstract}
Scene-level point cloud self-supervised learning (PC-SSL) has demonstrated potential in enhancing the generalization capability of 3D vision models. Despite the advances in the field through existing methods, the sample-independent modeling paradigm still poses significant limitations in terms of maintaining consistent semantic representations across scenes. This challenge hinders the construction of a unified and transferable semantic space. To address this issue, we propose a PC-SSL framework based on cross-sample semantic propagation (CSP), in which samples within a batch are serialized into continuous input and processed by a state-space model to enable semantic state propagation. This mechanism explicitly models the dynamic dependencies across samples in the state space, allowing the network to establish cross-sample semantic consistency in the latent space and  achieve global semantic alignment. Since serialization-based pretraining requires batch-level input organization, we further introduce an asymmetric semantic preservation distillation (SPD) during finetuning to achieve structural alignment of semantic transfer and eliminate inconsistencies caused by batch dependency. The proposed SPD ensures stable transfer of pretrained semantics through a heterogeneous input mechanism and a semantic feature alignment constraint. This enables the model to maintain structured semantic consistency and robustness under single-scene testing conditions. Extensive experiments on multiple benchmark datasets demonstrate that our method consistently outperforms state-of-the-art methods in both performance and semantic consistency.
\end{abstract}    
\section{Introduction}
\label{sec:intro}

\begin{figure}[t]
  \centering
   \includegraphics[width=0.95\linewidth]{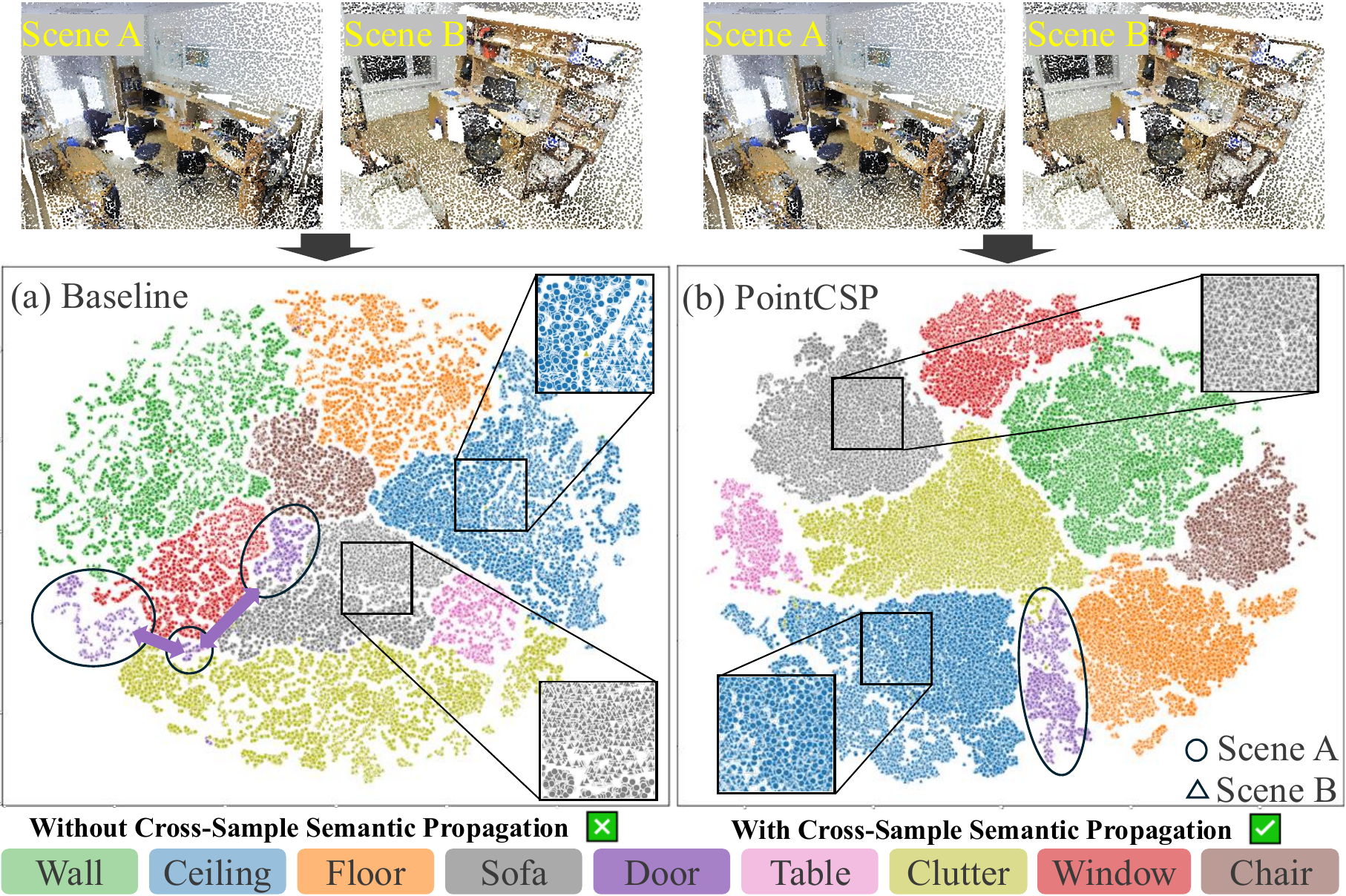}
   \caption{A comparison of t-SNE visualizations under identical semantic categories across different scenes. Scenes A and B share nine common semantic classes. (a) Without cross-sample semantic propagation, the features from both scenes are dispersed in the embedding space and fail to form consistent clusters. (b) Our method significantly reduces inter-scene discrepancies, yielding more compact and semantically aligned feature distributions.}
   \label{fig:csp}
\end{figure}

Point clouds provide a fundamental 3D representation of the real world, serving as the cornerstone for scene understanding \cite{zhu_living_2024,wang_masked_2025,wang2026mcgs,yu_facnet_2025}, autonomous perception \cite{chae_towards_2024,yin_fusion_2024,luo2024exploring,huang_l4dr_2025}, and embodied intelligence \cite{tian_pdfactor_2025,zhu_point_2024}.
Scene-level point cloud self-supervised learning (PC-SSL) has emerged as a powerful paradigm for learning transferable geometric and semantic representations from raw spatial data, thereby enabling a wide range of downstream 3D tasks.

Recent advances in PC-SSL have focused on enhancing geometric and semantic representation learning through multi-view aggregation, contextual reconstruction, and contrastive objectives.
Specifically, PointContrast \cite{xie_pointcontrast_2020} employs scene registration-based contrastive learning to construct positive and negative pairs between local and global views, thereby improving the transferability of 3D representations in an unsupervised manner. 
SSPL \cite{huang_spatio_2021} introduces a joint local-global contrastive pretraining strategy for large-scale scene point clouds, aggregating multi-scale features to strengthen global contextual awareness. 
MSP \cite{jiang_self_2023} further incorporates a masked shape prediction objective to reconstruct geometric structures, learning context-aware geometric features that benefit downstream tasks. 
Despite these advances, current PC-SSL methods predominantly emphasize local geometric feature aggregation and view-level contrastive alignment. 
However, due to their reliance on sample-independent modeling paradigms, the learned semantic representations remain inconsistent across scenes, as shown in \cref{fig:csp} (a), making it hard to establish a unified semantic space.
Such semantic inconsistency across scenes weakens the continuity of representation learning, hinders the establishment of a coherent global semantic topology and ultimately constrains cross-scene generalization.

To overcome the aforementioned limitations, it is essential to address several intrinsic challenges associated with scene-level point cloud, including the large spatial and semantic variability across scenes and the absence of explicit semantic continuity or dependency between samples.
The former leads to fragmented feature distributions, where features extracted from different scenes occupy disjoint regions in the latent space due to variations in spatial layout, object density, and contextual composition.
Such fragmentation prevents the model from learning consistent inter-scene semantics, resulting in domain-specific biases and limited cross-scene generalization.
The latter impedes the formation of inter-sample semantic dependencies, as scenes are modeled in isolation without contextual interaction, preventing the model from establishing coherent semantic transitions across samples and resulting in disconnected latent representations with incomplete global semantic alignment.
A straightforward way to mitigate these challenges is to enlarge the scale and diversity of training data to achieve a statistical scaling effect.
However, collecting and annotating large-scale scene-level point clouds is both costly and inefficient, making this approach impractical for scalable pretraining.
The discrepancy between data diversity and semantic consistency underscores the necessity of principled and scalable solutions that can establish coherent global semantics even under limited or imbalanced data conditions.

To address these challenges, we propose PointCSP, a self-supervised framework that integrates Cross-Sample Semantic Propagation (CSP) and Semantic Preservation Distillation (SPD) for scene-level point cloud learning. CSP captures semantic dependencies across samples through a continuous state-space modeling scheme, while SPD stabilizes semantic transfer across scenes via asymmetric teacher-student distillation.
Built upon a self-distillation framework inspired by DINO \cite{caron_emerging_2021}, CSP introduces a state-space modeling framework that explicitly captures semantic dependencies across samples, thereby overcoming the traditional assumption of sample independence.
In this framework, batch-level point cloud samples are serialized into a unified sequence and processed by a state-space model that recursively propagates semantic states along the sequence, enabling the network to capture long-range contextual dependencies and maintain inter-sample semantic consistency.
By embedding samples into a continuous semantic space, CSP simultaneously mitigates the large spatial and semantic variability across scenes through global feature alignment, while establishing semantic continuity across samples via recursive state propagation.
This design enables the model to capture both inter-scene coherence and inter-sample dependencies within a unified pretraining framework, facilitating a consistent and transferable global representation, as shown in \cref{fig:csp}(b), where features of identical semantic categories form more compact and well-aligned clusters.

However, since serialization-based pretraining relies on batch-level input organization, directly transferring the pretrained model to single-scene testing can lead to structural misalignment and semantic inconsistency. 
To overcome this limitation, SPD achieves structural alignment during semantic transfer and eliminates inconsistencies introduced by batch dependency.
Specifically, the teacher branch continues to process serialized sequences to preserve the global semantic topology established during pretraining, while the student branch operates on standard scene-level inputs and learns to align with the teacher through a semantic consistency constraint. 
This asymmetric design effectively bridges the gap between pretraining and downstream adaptation, ensuring stable semantic migration and robust generalization under limited-data conditions. 

Our contributions are summarized as follows:
\begin{itemize}
    \item We propose PointCSP, a self-supervised framework for learning point cloud representations that establishes a unified, transferable semantic space.
    \item We introduce a Cross-Sample Semantic Propagation (CSP) mechanism that models semantic dependencies across samples within a continuous state space, enabling consistent and transferable global representations.
    \item We further design a Semantic Preservation Distillation (SPD) mechanism that stabilizes semantic transfer during finetuning through asymmetric teacher-student alignment, effectively bridging the gap between pretraining and downstream adaptation.
    \item Extensive experiments on S3DIS, 3DSES, ScanObjectNN, ModelNet40 and ShapeNetPart demonstrate that PointCSP achieves state-of-the-art (SOTA) performance in cross-scene generalization and robust representation.
\end{itemize}
\section{Related Work}
\label{sec:related_work}

\begin{figure*}[t]
  \centering
   \includegraphics[width=0.95\linewidth]{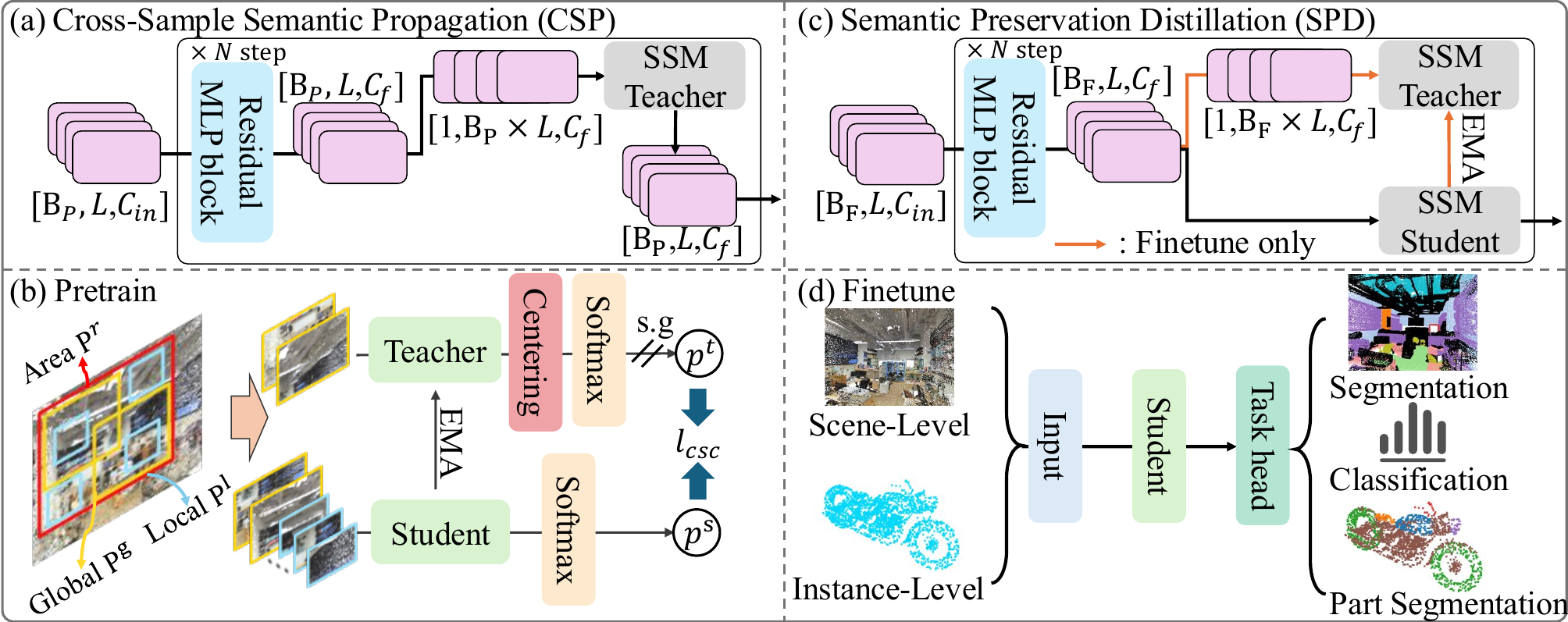}
   \caption{Overview of the proposed PointSCP.
(a) The CSP models cross-sample semantic dependencies within the state-space representation.
(b) The self-distillation pretraining framework, inspired by DINO, integrates CSP for unified semantic consistency learning.
(c) The SPD performs semantic preservation distillation during fine-tuning.
(d) The task-oriented downstream fine-tuning covers scene-level segmentation, instance-level classification, and part segmentation tasks.
The CSP is employed in (b), while SPD is applied in (d).
Let $B_{P}$ and $B_{F}$ denote the pretrain and finetune batch sizes, respectively; $C_{in}$ and $C_{f}$ represent the input and output feature channels.
}
   \label{fig:backbone}
\end{figure*}

\noindent  \textbf{Scene-level point cloud self-supervised learning.}
Recent advances in scene-level PC-SSL have evolved primarily along three complementary directions: multi-view aggregation, context reconstruction, and contrastive objectives.
Multi-view aggregation methods, such as Contrastive Scene Contexts \cite{hou_exploring_2021}, PointContrast \cite{xie_pointcontrast_2020}, SSPL \cite{huang_spatio_2021}, and DPCo \cite{li_closer_2022}, enhance spatial coherence by aligning representations across diverse scene views, temporal sequences, or sensor modalities, thereby learning view-invariant and geometry-aware features.
Context-reconstruction paradigms, including MSC \cite{wu_masked_2023}, Ponder \cite{huang_ponder_2023}, MM-3DScene \cite{xu_mm_2023}, PointDif \cite{zheng_point_2024}, Shape2Scene \cite{feng_shape2scene_2024}, MSP\cite{jiang_self_2023} and Masked Scene Modeling \cite{hermosilla_masked_2025}, focus on reconstructing masked, rendered, or diffusion-based contexts to recover holistic spatial semantics.
Meanwhile, contrastive-learning frameworks remain central to scene-level representation learning, as demonstrated by SegContrast \cite{nunes_segcontrast_2022}, GroupContrast \cite{wang_groupcontrast_2024}, OESSL \cite{wu_mitigating_2024}, Point-GCC \cite{fan_point_2024}, MPEC \cite{wang_masked_2025}, and Point-MoDE \cite{zha_point_2025}.
However, since these approaches are built upon independent sample modeling, they inherently lack semantic continuity across different scenes, thereby impeding the establishment of a globally consistent and transferable semantic space.

\noindent  \textbf{Point cloud learning based on the state space model (SSMs).}
SSMs have emerged a paradigm for point cloud learning, with representative methods including Point Mamba–based architectures such as Manba3D \cite{han_mamba3d_2024}, PointMamba \cite{liang_pointmamba_2024}, PointTrMamba \cite{wang_pointramba_2024}, PCM \cite{zhang_point_2025}, MambaMOS \cite{zeng_mambamos_2024}, FEAST-Mamba \cite{li_feast_2025}, Pamba \cite{li_pamba_2025}, PointMC \cite{yu2026pointmc}, and SI-Mamba \cite{bahri_spectral_2025}.
The Point Mamba family adapts the selective state-space mechanism of Mamba \cite{gu_mamba_2024} or Mamba2 \cite{dao_mamba2_2024} to unordered and irregular 3D data through spatial serialization \cite{hilbert_stetige_1935, morton_computer_1966}, local-to-global gating, and efficient token interaction, thereby enabling robust long-range context modeling with linear complexity.
Collectively, these works underscore the potential of SSMs in capturing intricate geometric dependencies, improving scalability, and providing an efficient alternative backbone for diverse 3D tasks.
\section{Methods}
\label{sec:methods}

\subsection{Cross-Sample Semantic Propagation (CSP)}
\noindent  \textbf{SSM Based Context Propagation}
The goal of Cross-Sample Semantic Propagation (CSP) is to establish semantic continuity across samples during scene-level point cloud pretraining.
Moving beyond the conventional assumption of sample independence, CSP explicitly models inter-sample semantic dependencies within a continuous state space.
The overall architecture is shown in \cref{fig:backbone}(a).
Given the feature sequence $\mathbf{F}$ produced by the residual MLP block, we reshape it into a unified long sequence $\mathbf{F}'$ before feeding it into the SSM module.
The SSM then recursively propagates hidden semantic states along the serialized dimension, with each state encoding both intra-scene context and inter-sample semantic correlations.
This formulation elevates pretraining from local instance alignment to global contextual reasoning, producing feature embeddings that preserve higher-order semantic relationships across scenes.

Formally, given a batch of $B$ point cloud samples, each sample is represented as a feature sequence 
\begin{equation}
\mathbf{F}_i = [\mathbf{f}_{i,1}, \mathbf{f}_{i,2}, \ldots, \mathbf{f}_{i,L}] \in \mathbb{R}^{L\times C}, 
\end{equation}
where $L$ denotes the number of points and $C$ the dimension of the feature.
To enable inter-sample propagation, we reshape the batch into a unified serialized sequence:
\begin{equation}
\mathbf{F}' = [\mathbf{f}_{1,1}, \ldots, \mathbf{f}_{1,L}, \mathbf{f}_{2,1}, \ldots, \mathbf{f}_{B,L}] \in \mathbb{R}^{(B\times L)\times C},
\end{equation}
and feed it into the SSM block. The SSM maintains a hidden state $\mathbf{h}_t \in \mathbb{R}^{C}$ that evolves over the serialized dimension according to the recursive dynamics:
\begin{equation}
\mathbf{h}_t = f_{\theta}(\mathbf{A}\mathbf{h}_{t-1} + \mathbf{B}\mathbf{f}'_t), \quad \mathbf{y}_t = \mathbf{C}\mathbf{h}_t,
\end{equation}
where $\mathbf{A}, \mathbf{B}, \mathbf{C}$ are learnable projection matrices, and $f_\theta(\cdot)$ denotes the nonlinear transition operator parameterized by the SSM block.
This formulation allows the hidden state $\mathbf{h}_t$ to integrate contextual cues from both the current point and all preceding samples, forming a continuous semantic flow along the serialized dimension.
The resulting latent representation encodes both intra-scene contextual dependencies and inter-sample semantic continuity, thereby bridging discrete scene embeddings into a unified and globally consistent semantic space.
Through the continuous state-space evolution process, the proposed CSP mechanism enables cross-sample knowledge propagation and effectively extends the semantic receptive field beyond individual samples, further reinforcing global semantic alignment.

\noindent  \textbf{Cross-Sample Self-Distillation Based on DINO.}
The overall architecture is illustrated in \cref{fig:backbone} (b).
To better exploit their structural characteristics, we design a multi-region augmentation strategy that enhances both spatial diversity and semantic coverage during pretraining.
Given an input point cloud $P^{s}$, a candidate region $P^{r}$ covering approximately 60\% of the full point cloud is first selected and normalized. From $P^{r}$, we randomly sample $n$ global subregions ${P_i^{g}}, i \in [1,n]$ with broader spatial coverage ratios of about $[40\%,80\%]$ of $P^{r}$, and $m$ local subregions ${P_j^{l}}, j \in [1,m]$ with finer spatial coverage ratios of about $[10\%,30\%]$ of $P^{r}$. These global and local subregions serve as complementary input views, allowing the model to capture hierarchical semantics and multi-scale contextual dependencies during pretraining.
The $P^{g}$ and $P^{l}$ are fed into the student network $N_\omega(\cdot)$, while only $P^{g}$ are forwarded through the teacher network $N_\varphi(\cdot)$.
Both networks share identical architectures but differ in their parameter update mechanisms: the student is optimized via back-propagation, whereas the teacher is updated as an exponential moving average (EMA) of the student to provide stable supervision.

Formally, we denote the teacher and student outputs as $\mathbf{p}^t$ and $\mathbf{p}^s$. To encourage local-global semantic consistency, each student view output is matched against the global teacher output, 
yielding the semantic consistency loss,
\begin{equation}
\mathcal{L}_{\text{CSC}} = 
-\frac{1}{n (m+n)} 
\sum_{i=1}^{n} \sum_{j=1}^{(m+n)} 
\mathbf{p}_i^t \cdot \log \mathbf{p}_j^s .
\end{equation}
During training, the teacher parameters are updated as,
\begin{equation}
N_\varphi \leftarrow \gamma \, N_\varphi + (1 - \gamma)\, N_\omega ,
\end{equation}
where $\gamma \in [0,1)$ is the EMA momentum coefficient.

\subsection{Semantic Preservation Distillation (SPD)}

Although strong semantic continuity has been established during pretraining through the cross-sample state propagation mechanism, the serialized batch-level organization inevitably introduces structural dependency across samples.
Consequently, when transferring the pretrained model to downstream tasks, the absence of such batch-level contextual support often leads to semantic structure degradation and representation drift, especially under sample-limited finetuning conditions.
To address this issue, we introduce an SPD framework, which achieves alignment during semantic transfer and mitigates inconsistencies caused by batch dependency.
By maintaining the global semantic topology established in pretraining, SPD enables a smooth transition of structural semantics and preserves the continuity of learned representations throughout finetuning stage.

As illustrated in \cref{fig:backbone} (c), the teacher network aggregates batch-level features to capture the global semantic distribution, whereas the student network learns to internalize and reconstruct these semantic structures under standard input conditions. 
Formally, given an input point cloud batch, the teacher and student networks produce feature embeddings $\mathbf{F}^t$ and $\mathbf{F}^s$.
To preserve global semantic structure, we enforce a consistency constraint between the two representations features using a mean squared error loss:
\begin{equation}
\mathcal{L}_{\text{SPD}} = 
\frac{1}{N} \sum_{i=1}^{N}
\|\mathbf{f}_i^s - \mathbf{f}_i^t\|_2^2,
\end{equation}
where $N$ is the number of points in the point cloud.

Although the cross-sample SSM module has already endowed the model with semantic propagation ability during pretraining, SPD further updates the teacher parameters via  EMA strategy, forming temporally smoothed semantic anchors. 
This design enables the teacher to maintain a stable semantic manifold while the student parameters are rapidly optimized for downstream objectives.
During inference, only the student network is employed for feature extraction, ensuring efficient deployment. 
Such an asymmetric training mechanism allows the student network to reproduce the global semantic topology established during pretraining, even under context-limited inference scenarios. 
Consequently, SPD facilitates a smooth transfer of semantic structures from pretraining to inference, ensuring discriminative completeness, structural stability, and semantic consistency across the entire training-deployment pipeline.

\subsection{Optimization Objective}

To further enhance the geometric reversibility of intermediate features across different semantic depths, we extend the geometric consistency constraint into a multi-level formulation. 
Specifically, from multiple encoder layers, a subset of feature vectors $\{f_k^{(l)}\}$ is randomly sampled at each layer $l$, and decoded into their corresponding 3D coordinates through a lightweight coordinate decoder $D_\phi^{(l)}(\cdot)$:
\begin{equation}
\hat{\mathbf{x}}_k^{(l)} = D_\phi^{(l)}(f_k^{(l)}).
\end{equation}
Each layer-specific geometric reconstruction loss is defined:
\begin{equation}
\mathcal{L}_{\text{geo}}^{(l)} =
\frac{1}{K_s^{(l)}}\sum_{k=1}^{K_s^{(l)}}
\|\hat{\mathbf{x}}_k^{(l)} - \mathbf{x}_k^{(l)}\|_2^2,
\end{equation}
where $K_s^{(l)}$ denotes the number of randomly sampled features from the $l$-th layer. This layer-wise constraint encourages both shallow and deep features to preserve geometric consistency with the original 3D structure. The overall geometric consistency loss is then aggregated over multiple layers as:
\begin{equation}
\mathcal{L}_{\text{geo}} = 
\sum_{l=1}^{L} \alpha_l\, \mathcal{L}_{\text{geo}}^{(l)},
\end{equation}
where $\alpha_l$ is the weight assigned to the $l$-th layer, and $L$ is the number of supervised layers. This multi-level geometric constraint explicitly regularizes the representation hierarchy, enforcing geometric reversibility at different semantic depths and leading to structurally coherent and semantically consistent 3D feature learning.

Finally, the pretraining objective combines the semantic consistency and multi-level geometric constraints:
\begin{equation}
\mathcal{L}_{\text{pretrain}} 
= \mathcal{L}_{\text{CSC}} 
+ \lambda_{\text{geo}}\, \mathcal{L}_{\text{geo}}.
\end{equation}
the finetuning objective integrates the semantic preservation distillation with task-specific losses:
\begin{equation}
\mathcal{L}_{\text{fine-tune}} 
= \mathcal{L}_{\text{task}} 
+ \lambda_{\text{SPD}}\, \mathcal{L}_{\text{SPD}}
+ \lambda_{\text{geo}}\, \mathcal{L}_{\text{geo}},
\end{equation}
where $\mathcal{L}_{\text{task}}$ is task based loss function, $\lambda_{\text{geo}}$ and $\lambda_{\text{SPD}}$ are hyperparameters balancing the respective loss terms.
\section{Experiments}
\label{sec:experiments}

\begin{table}[t]
    \caption{Semantic segmentation mIoU results (\%) on the S3DIS dataset are evaluated on Area5 and 6-fold cross-validation.}
    \label{tab:S3DIS}
    \centering
    \setlength{\tabcolsep}{7pt}
    \begin{tabular}{l c c c}
        \toprule
        Method                                  &   Ref.        & Area5             & 6-fold      \\
        \midrule
        PointDif\cite{zheng_point_2024}         &   CVPR'24     & 70.0              & /           \\
        PTV3\cite{wu_point_2024}                &   CVPR'24     & 73.4              & 77.7        \\
        HPENet\cite{zou_improved_2024}          &   AAAI'24     & 72.7              & 78.7        \\
        PDNet-XXL\cite{yin_point_2024}          &   AAAI'24     & 72.3              & 78.3        \\ 
        Point-MoDE\cite{zha_point_2025}         &   IJCAI'25    & 70.9              & /           \\
        DeepLA-120\cite{zeng_deepla-net_2025}   &	CVPR'25     & 75.7              & 79.8        \\
        CamPoint\cite{zhang_campoint_2025}      &   CVPR'25     & \underline{83.3}  & /           \\
        Sonata\cite{wu_sonata_2025}             &   CVPR'25     & 76.0              & \underline{82.3}        \\
        VDG\cite{han_all_2025}                  &   ICCV'25     & 71.5              & 73.2        \\
        Point-PQAE\cite{zhang_towards_2025}     &   ICCV'25     & 61.4              & /           \\
        \midrule
        PointCSP                                &   /           &\textbf{88.2}      &\textbf{93.1}\\
        \bottomrule
    \end{tabular}
\end{table}%

\begin{figure*}[t]
\centering
\includegraphics[width=0.925\linewidth]{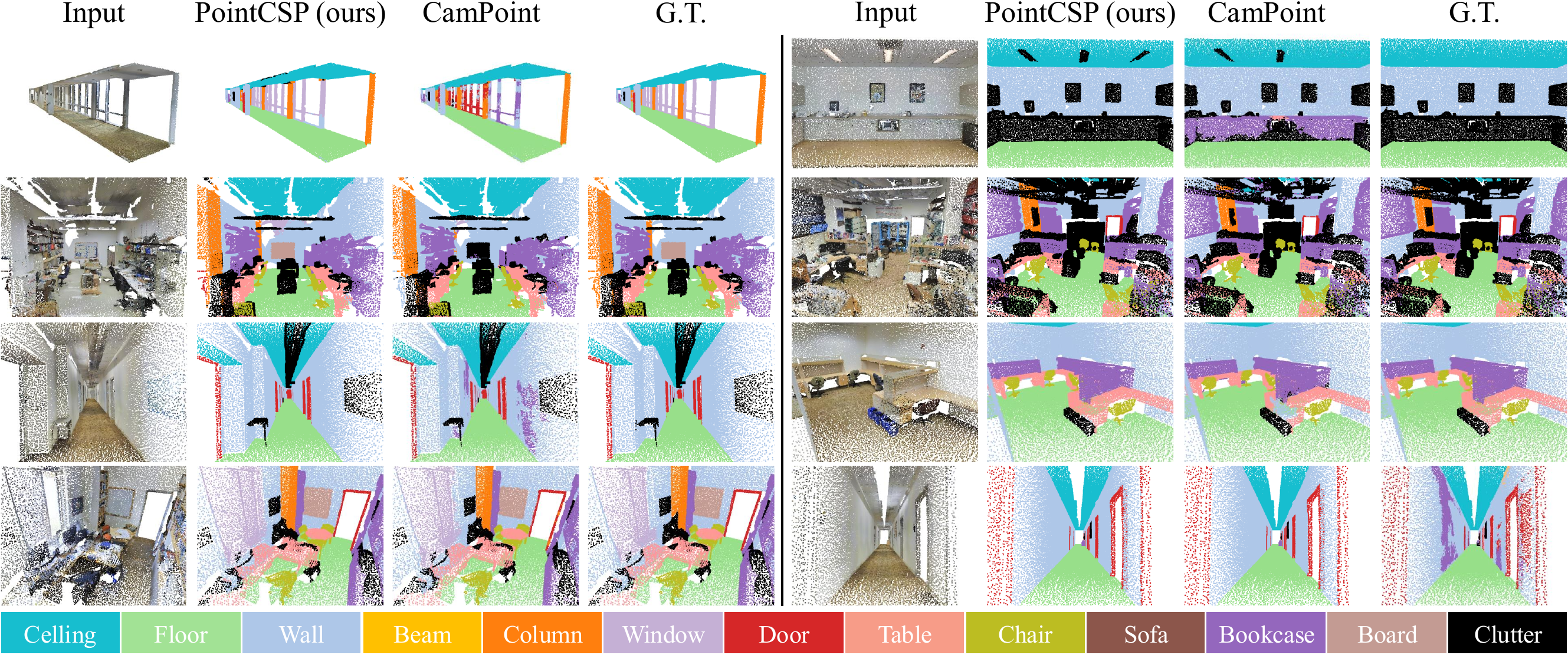} 
\caption{Visualization of large-scale object-level semantic segmentation results on the S3DIS Area-5 dataset, showing input point clouds, ground truth, and predicted results from PointCSP and the current SOTA method CamPoint.}
\label{pic:S3DIS_VIS}
\end{figure*}

\subsection{Implementation Details}

We pretrain our model on the widely used ScanNetV2 dataset \cite{dai_scannet_2017}, which contains 1513 indoor scenes captured by RGB-D sensors.
During pretraining, each point cloud instance is divided into two global and eight local regions to capture multi-scale contextual information. A hierarchical downsampling strategy is adopted: voxel-based uniform sampling \cite{chen_decoupled_2023} is applied to $P^{g}$ to maintain balanced spatial coverage, while Farthest Point Sampling is used for $P^{l}$ to preserve representative geometric details. In the final stage, both $P^{g}$ and $P^{l}$ subsets are uniformly reduced to 1024 points, ensuring consistent feature dimensionality and a unified representation scale across samples.
For the backbone \cite{zhang_campoint_2025}, we employ an enhanced version of Mamba2, referred to as Gated DeltaNet \cite{yang_gated_2025}.
In contrast to conventional SSM-based methods that rely on explicit spatial ordering, each SSM block in our design processes randomly shuffled point cloud tokens without predefined sorting. This allows the model to learn sequential semantic dependencies independent of spatial priors, reinforcing the robustness of semantic propagation in unordered point sets.
The network is optimized using AdamW \cite{loshchilov_decoupled_2017} with a cosine learning rate schedule and an initial warm-up phase to stabilize early training dynamics.
During evaluation, we follow an input-constrained inference setting, where the model processes one scene or instance per forward pass. This setup prevents contextual information leakage across scenes, ensuring a fair assessment of scene-level generalization.

\subsection{Finetuning in Scene-level Downstream Tasks}
\noindent  \textbf{Semantic Segmentation on S3DIS Dataset.} 
We first evaluate our method on the S3DIS \cite{armeni_3d_2016} dataset, which consists of six distinct indoor areas, each comprising multiple rooms with diverse spatial layouts, scene complexities, and object distributions. Two widely used evaluation protocols are adopted, contain Area 5 and 6-fold cross-validation. The mean Intersection over Union (mIoU) serves as the principal evaluation metric for quantifying accuracy.

As shown in \cref{tab:S3DIS}, our method achieves 88.2\% mIoU on the Area 5 split and 93.1\% mIoU under the 6-fold cross-validation protocol, establishing a new SOTA on the S3DIS benchmark.
This superior performance arises from the consistent use of scene-level data during both pretraining and finetuning, which mitigates the semantic inconsistency typically caused by instance-to-scene transfer.
By leveraging large-scale contextual learning across diverse indoor environments, the proposed framework effectively captures cross-scene semantic dependencies and constructs globally continuous structural representations that remain stable during adaptation.
Such semantic continuity promotes coherent feature organization within the embedding space, thereby enhancing both discriminative power and generalization in complex 3D scene understanding tasks.

\begin{table}[t]
    \caption{Quantitative comparison of edge-strip segmentation with state-of-the-art methods on S3DIS, reporting mIoU (\%) and F1-scores (\%) under different neighborhood radii (R).}
    \label{tab:S3DIS_boundaries}
    \centering
    \resizebox{1\linewidth}{!}{
    \setlength{\tabcolsep}{2pt}
    \begin{tabular}{l c c c c c c c}
        \toprule
        \multirow{2}{*}{Method} &   Overall     & \multicolumn{2}{c}{R=0.01}  & \multicolumn{2}{c}{R=0.02} & \multicolumn{2}{c}{R=0.04}     \\ 
        ~                       &   mIoU    & mIoU         & F1             & mIoU         & F1             & mIoU         & F1        \\
        \midrule
        CamPoint \cite{zhang_campoint_2025}                &   83.3        & 74.9             & 84.7               & 79.2             & 89.2               & 81.8             & 92.4   \\
        \midrule
        PointCSP                     &   \textbf{88.2}        & \textbf{80.0}             & \textbf{90.0}               & \textbf{84.8}             & \textbf{93.2}               & \textbf{88.0}             & \textbf{95.6}   \\
        \bottomrule
    \end{tabular}}
\end{table}%

To further evaluate the robustness of our method, we perform an edge-strip analysis on the S3DIS dataset with varying neighborhood radii.
As shown in \cref{tab:S3DIS_boundaries}, the proposed model consistently outperforms CamPoint across all settings, demonstrating superior boundary awareness and multi-scale stability.
The observed improvements stem from the collaborative effect of CSP and SPD, which jointly maintain semantic consistency throughout pretraining and finetuning.
During pretraining, CSP enforces local–global alignment and cross-sample state propagation, establishing a continuous semantic topology that links spatially discrete object regions.
In finetuning, CSP maintains cross-sample semantic continuity by recursively propagating latent states through the serialized batch sequence.
This mechanism enables the model to aggregate contextual information beyond individual scenes, allowing it to capture long-range semantic dependencies and boundary transitions often fragmented in conventional instance-level training.

\begin{table*}[t]
    \caption{Segmentation results on the 3DSES Silver dataset. Comparison of different methods under pseudo labels (P) and real labels (R). Metrics include OA (\%), mIoU (\%), and per-class IoU (\%). Our PointCSP consistently achieves the highest scores across both settings.}
    \label{tab:3DSES}
    \centering
    \resizebox{1\linewidth}{!}{
    \begin{tabular}{l c | c c | c c c c c c c c c c c c}
        \toprule
        Method   &   Labels  & OA       & mIoU  & Column & Cover & Door & Exit sign & Heater & Lamp & Railing & Slab & Stair & Wall & Window & Clutter \\ 
        \midrule
        PointNeXt-S\cite{qian_pointnext_2022}  &  P     & 92.58   &62.91 & 58.44  & 96.55    & 69.81& 0.00      & 33.96  & 67.00 & 38.90   & 93.86& 83.48 & 88.12& 51.25  & 73.60  \\
        Swin3D\cite{yang_swin3d_2025}     &    P     & 93.46   &65.70 & 52.31  & 95.82    & 89.01& 11.79     & 65.29  & 55.28& 64.17   & 82.06& 34.32 & 92.44& 54.00  & 91.92  \\
        \midrule        
        PointCSP   &    P       & \textbf{96.41} &\textbf{69.07} & 68.27& 97.51& 92.49& 5.57& 80.29 & 60.34 & 27.73& 93.47& 60.30& 94.65 & 55.38& 92.80  \\
        \midrule
        PointNeXt-S\cite{qian_pointnext_2022}  &  R     & 94.63   &64.99 & 0.00   & 97.07    & 76.66& 0.00      & 38.73  & 78.11  &65.26  & 94.85 & 86.97 & 90.84 & 67.08  & 84.35  \\
        Swin3D\cite{yang_swin3d_2025}     &    R     & 90.47   &57.08 & 5.40   & 94.35    & 83.06& 9.30      & 75.27  & 44.04  &37.63  & 84.08 & 38.69 & 85.34 & 54.99  & 72.83  \\
        \midrule        
        PointCSP   &    R    & \textbf{95.78} &\textbf{65.13} & 67.16& 97.46& 93.24& 6.24& 90.04& 63.06& 13.12& 93.49& 23.70& 94.26& 52.17& 87.57  \\
        \bottomrule
    \end{tabular}}
\end{table*}%

\begin{table}[t]
    \caption{Classification OA (\%) and mAcc(\%) on ModelNet40 and ScanObjectNN. The ScanObjectNN results are reported on the most challenging PB\_T50\_RS variant.}
    \label{tab:ScanObjectNN}
    \centering
    \resizebox{1\linewidth}{!}{
    \begin{tabular}{l c c c c c}
    \toprule
        \multirow{2}{*}{Method}                 & \multirow{2}{*}{Ref.} & \multicolumn{2}{c}{ScanObjectNN}      & \multicolumn{2}{c}{ModelNet40}        \\
                    ~                                   & ~             &    OA             &mAcc               &       OA          &mAcc               \\
        \midrule
        LinNet\cite{deng_linnet_2024}                   & NeurIPS 24    & 88.2              & 86.6              & 93.6             & 91.0               \\ 
        KPConvX-L\cite{thomas_kpconvx_2024}             & CVPR 24       & 89.3              & 88.1              & /                & /                  \\
        Mamba3D\cite{han_mamba3d_2024}                  & ACMM 24       & 91.8              & /                 & 93.4             & /                  \\ 
        PCM \cite{zhang_point_2025}                     & AAAI 25       & 88.1              & 86.6              & 93.4             & 90.7               \\ 
        CamPoint\cite{zhang_campoint_2025}              & CVPR 25       & \underline{92.1}  & \underline{91.1}  & 93.6             & \underline{91.3}   \\
        Point-MoDE\cite{zha_point_2025}                 & IJCAI 25      & 88.6              & /                 & \underline{93.6} & /                  \\    
        SI-Mamba\cite{bahri_spectral_2025}              & CVPR 25       & 87.3              & /                 & 92.7             & /                  \\  
        PointSD\cite{chen_harnessing_2025}              & ICCV 25       & 90.1              & /                 & 93.7             & /                  \\  
        Point-PQAE\cite{zhang_towards_2025}             & ICCV 25       & 83.3              & /                 & 92.8             & /                  \\  
        \midrule
        PointCSP                                        & ~             & \textbf{92.8}     & \textbf{91.7}     & \textbf{93.9}    & \textbf{91.5}      \\ 
        \bottomrule
    \end{tabular}}
\end{table}

\begin{figure}[t]
\centering
\includegraphics[width=\linewidth]{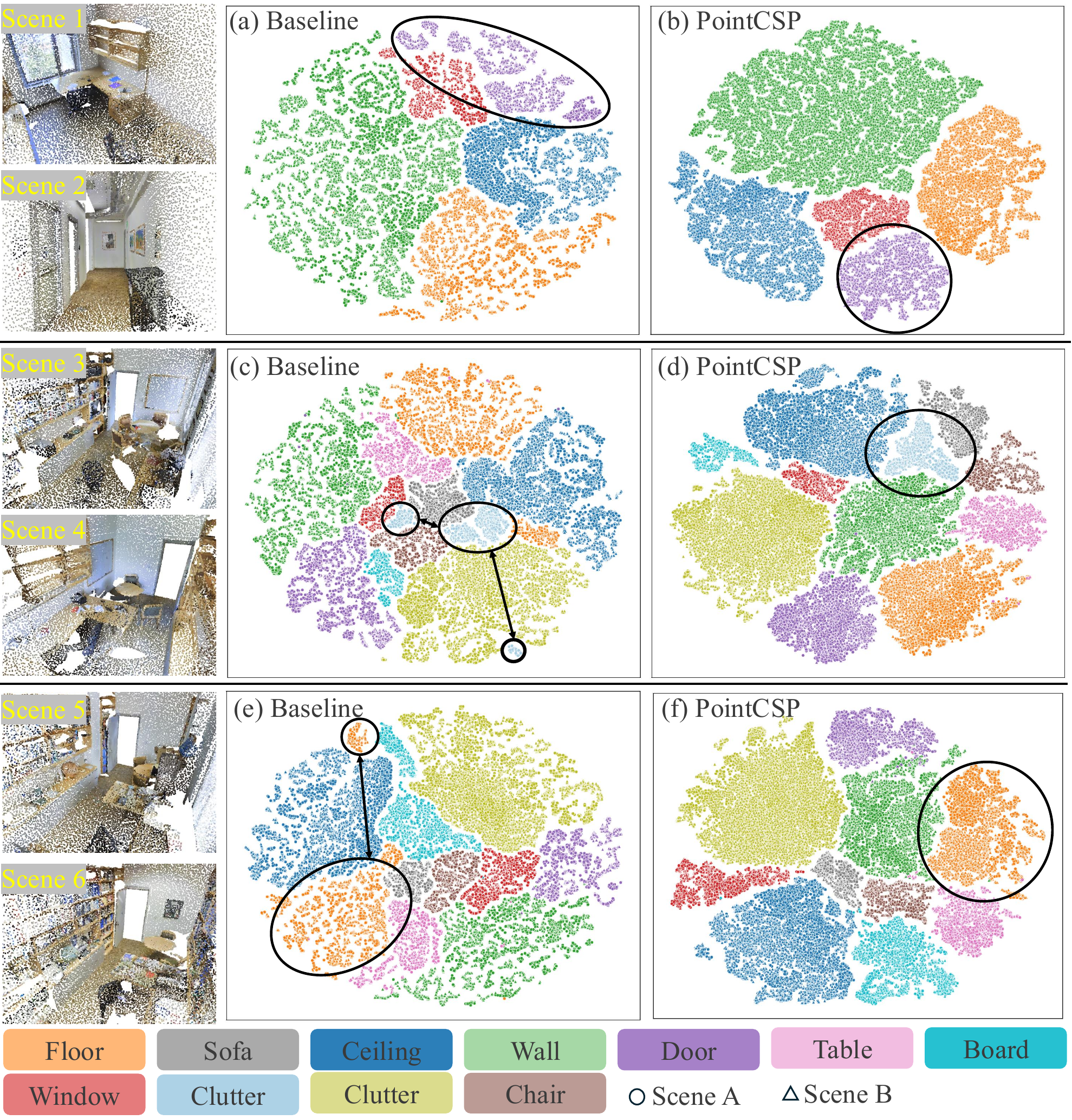} 
\caption{t-SNE visualizations of features across three scene groups. The first group contains an office and a corridor, while the second and third groups both correspond to office scenes.}
\label{pic:morecsp}
\end{figure}

The visualization in \cref{pic:S3DIS_VIS} illustrates the qualitative improvements achieved by PointCSP.
Compared with the current SOTA method CamPoint, PointCSP exhibits more accurate boundary delineation and clearer semantic discrimination in the segmentation of objects, effectively reducing category confusion in complex indoor layouts.
Furthermore, as shown in \cref{pic:morecsp}, we present t-distributed stochastic neighbor embedding (t-SNE) \cite{maaten_visualizing_2008} visualizations of features with identical semantic categories across three groups.
Compared with the baseline, our method achieves better feature aggregation and semantic compactness, resulting in more coherent aligned clusters across diverse scenes.

\noindent  \textbf{Semantic Segmentation on 3DSES Dataset.} 
To rigorously assess the robustness and cross-domain generalization capability of PointCSP, the PointCSP is further evaluated on the 3DSES Silver \cite{visapp25} dataset.
In comparison with conventional indoor benchmarks such as S3DIS and ScanNetV2, which are reconstructed from RGB-D imagery, 3DSES Silver is captured via Terrestrial Laser Scanning (TLS), yielding subcentimeter-level dense point clouds with exceptional geometric fidelity.
The dataset comprises 30 indoor scenes, each containing approximately 7 million points, providing large-scale, high-density geometric observations that capture complex indoor structures.
Such high-resolution data preserve intricate architectural details and fine-grained surface variations, thereby substantially increasing the geometric and semantic complexity of the segmentation task.
The dataset encompasses a further 12 BIM-oriented semantic categories. These categories cover both structural and functional elements within indoor environments.

As summarized in \cref{tab:3DSES}, PointCSP consistently outperforms existing baselines in both pseudo-label and real-label settings.
Specifically, it achieves 96.41\% OA with pseudo labels and 95.78\% OA with real labels, surpassing vanilla approaches by a notable margin.
These results highlight the robustness of the model and its ability to maintain semantic continuity and structural stability in dense TLS scans, thereby showcasing its notable adaptability to complex, real-world indoor environments.

\subsection{Finetuning in Instance-level Downstream Tasks}
To further examine the generalization capability of the proposed method, we also conduct finetuning on instance-level downstream tasks, including shape classification and part segmentation. 
Although model has been pretrained on scene-level data, it is imperative to evaluate it on instance-level datasets in order to verify whether the learned semantic continuity can be transferred beyond scene contexts and adapted to isolated object representations. 
This setting facilitates a more rigorous evaluation of semantic scalability. In order to ensure a fair comparison with prior work, no voting or ensemble mechanism is employed during testing.

\noindent  \textbf{Shape Classification on ModelNet40 and ScanObjectNN Dataset.} 
We conduct experiments on two widely benchmarks, ModelNet40 \cite{wu_3d_2015} and ScanObjectNN \cite{uy_revisiting_2019}. The ModelNet40 dataset contains 12311 CAD models from 40 object categories, providing a clean synthetic environment for assessing category-level discrimination. In contrast, ScanObjectNN is a real-world dataset comprising approximately 15000 object instances with cluttered backgrounds, partial occlusions, and noise, where the PB\_T50\_RS variant serves as the most challenging benchmark.

As shown in \cref{tab:ScanObjectNN}, our method achieves 93.9\% classification overall accuracy (OA) on ModelNet40 and 92.8\% on ScanObjectNN, surpassing existing approaches and setting new SOTA results. 
The substantial improvement on ScanObjectNN, in particular, demonstrates the effectiveness of scene-level pretraining in transferring contextual knowledge to instance-level recognition under realistic conditions. 
By learning from large-scale scene collections, the model develops a structured and semantically coherent representation space that enhances category discrimination even when domain statistics and geometric completeness differ significantly between pretraining and finetuning. 

\noindent  \textbf{Part Segmentation on ShapeNetPart Dataset.} 
We also evaluate PointSCP on the ShapeNetPart \cite{yi_scalable_2016} dataset, which contains 16881 shapes from 16 object categories with 50 annotated part labels. This dataset serves as a benchmark for evaluating the capability of models to integrate fine-grained geometric details with coherent global semantic structures within instance-level representations.

As shown in \cref{tab:ShapeNetPart}, our method achieves a SOTA Ins. mIoU of 86.8\%.
While previous instance-level approaches primarily focus on optimizing local surface geometry, our model benefits from the global semantic priors established during scene-level pretraining, enabling it to generate more spatially coherent and structurally consistent part segmentation results.
The strong performance demonstrates that semantic continuity learned from large-scale scene contexts effectively transfers to fine-grained structural understanding, highlighting the scalability of the proposed framework across hierarchical semantic levels, ranging from holistic scene comprehension to detailed object decomposition.

\begin{table}[t]
    \caption{Part segmentation results on ShapeNetPart. The mIoU (\%) for all classes (Cls.) and instances (Ins.) are reported.}
    \label{tab:ShapeNetPart}
    \centering
    \resizebox{1\linewidth}{!}{
    \begin{tabular}{l c c c c}
    \toprule
        Method                                          &Ref.           & Ins. mIoU         & Cls. mIoU       \\
        \midrule
        PointGPT-S\cite{chen_pointgpt_2023}             & NeurIPS 23    & 86.2              & 84.1            \\ 
        PointMamba\cite{liang_pointmamba_2024}          & NeurIPS 24    & 86.2              & 84.4            \\ 
        Mamba3D\cite{han_mamba3d_2024}                  & ACMM 24       & 85.7              & 83.6            \\ 
        Point-MoDE\cite{zha_point_2025}                 & IJCAI 25      & \underline{86.5}  & \underline{84.6}            \\    
        SI-Mamba\cite{bahri_spectral_2025}              & CVPR 25       & 85.9              & /               \\  
        PointSD\cite{chen_harnessing_2025}              & ICCV 25       & 86.1              & 84.5            \\  
        Point-PQAE\cite{zhang_towards_2025}             & ICCV 25       & 86.1              & 84.6            \\  
        \midrule    
        PointCSP                                        & ~             & \textbf{86.8}     & \textbf{84.9}   \\ 
        \bottomrule
    \end{tabular}}
\end{table}

\subsection{Ablation Study}
Ablation experiments are conducted on the S3DIS dataset to evaluate the effectiveness of each component in PointCSP. 

\noindent  \textbf{Analysis of Each Component.} 
As shown in \cref{tab:Ablation_Component}, starting from the baseline containing only the backbone without semantic propagation or distillation, our model achieves 83.8\% mIoU.
Introducing the SPD module raises performance to 85.0\% mIoU, indicating that even without pretraining, semantic preservation distillation effectively stabilizes feature adaptation and enhances structural consistency during finetuning.
As reported in \cref{tab:Ablation_spd}, integrating SPD into CamPoint further verifies this effect.
We then evaluate the pretrained model fine-tuned with standard inputs, where the absence of cross-sample guidance leads to degraded generalization.
When both CSP and SPD are enabled, the model reaches 88.2\% mIoU, forming the complete PointCSP framework.
These results demonstrate that CSP and SPD act synergistically: CSP promotes semantic continuity and global alignment during pretraining, while SPD preserves this structure throughout finetuning, yielding consistent improvements in accuracy and semantic stability.

\begin{table}[t]
    \caption{The efficacy of each component in PointCSP.}
    \label{tab:Ablation_Component}
    \centering
    \begin{tabular}{ c c c c c c}
        \toprule
        Baseline                    & CSP                       & SPD                   & mIoU          & mAcc          & OA          \\
        \midrule    
        \checkmark                  &                           &                       & 84.7          & 87.7          & 96.8        \\
        \checkmark                  &                           &\checkmark             & 86.8          & 89.1          & 97.8          \\
        \checkmark                  &\checkmark                 &                       & 87.8          & 90.3          & 97.5          \\
        \checkmark                  &\checkmark                 &\checkmark             & \textbf{88.2} & \textbf{90.6} & \textbf{98.1}\\
        \bottomrule
    \end{tabular}
\end{table}

\begin{table}[t]
    \caption{Application of SPD in CamPoint.}
    \label{tab:Ablation_spd}
    \centering
    \setlength{\tabcolsep}{2.5pt}
    \begin{tabular}{l c c c}
        \toprule
        Method          &mIoU   &mAcc   &OA  \\
        \midrule    
        CamPoint        &83.3   &86.9   &96.0    \\
        CamPoint+SPD    &85.3(\textbf{+2.0})   &87.9(\textbf{+1.0})   &96.2(\textbf{+0.2})    \\
        \bottomrule
    \end{tabular}
\end{table}%


\begin{figure}[t]
\centering
\includegraphics[width=0.80\linewidth]{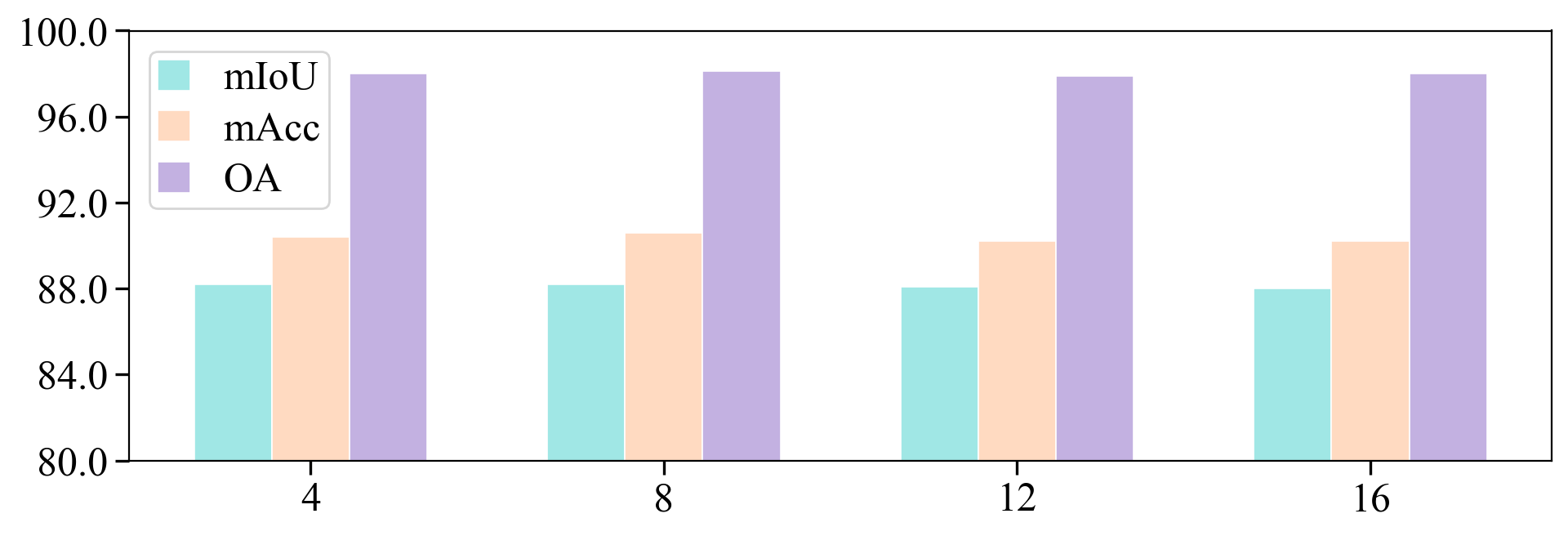} 
\caption{Ablation experiments on batch size during finetuning.}
\label{pic:metric}
\end{figure}

\noindent  \textbf{Batch Size Sensitivity during finetuning.} 
We analyze the sensitivity of PointCSP to varying batch sizes during finetuning to verify whether the model retains dependency on batch-level contextual information established during pretraining.
As shown in \cref{pic:metric}, the performance remains highly stable across different batch sizes, suggesting that the finetuning process is decoupled from the batch-size dependency introduced in serialized pretraining.
This observation demonstrates that the SPD mechanism maintains semantic alignment at the instance level, enabling effective adaptation without requiring batch-level contextual cues.

\section{Conclusion}
\label{sec:conclusion}

This study presents PointCSP, which addresses the challenge of constructing a unified and transferable semantic space. 
We propose CSP to model cross-sample dynamic dependencies and achieve global semantic alignment, and design SPD to stabilize semantic transfer through an asymmetric teacher-student structure, bridging pretraining and downstream adaptation.
Extensive evaluations show that PointCSP achieves SOTA results in scene segmentation, object classification, and part segmentation.

\noindent  \textbf{Acknowledgments.} 
This work was supported by the Science and Technology Development Fund of Macau Project 0096/2023/RIA2, 0123/2022/A3, 0044/2024/AGJ, 0140/2024/AGJ, the Chinese National Natural Science Foundation Projects 62406320, and the AI Super Computing Platform of Macau University of Science and Technology.

{
    \small
    \bibliographystyle{ieeenat_fullname}
    \bibliography{main}
}


\end{document}